\documentclass{llncs}
\usepackage{makeidx}  

\usepackage[T1]{fontenc}
\usepackage[utf8]{inputenc}
\usepackage{graphicx}
\usepackage{hyperref}
\usepackage{times}

\graphicspath{{./images/}}

\newcommand{\gf}[1]{\texttt{#1}}
\newcommand{\owl}[1]{\emph{#1}}

\begin{document}
\frontmatter          
\pagestyle{headings}  
\mainmatter              
\title{A Multilingual Semantic Wiki Based on Attempto Controlled English and
Grammatical Framework}
%
\titlerunning{A Multilingual Semantic Wiki Based on Attempto Controlled English
and Grammatical Framework}
%
%
\author{Kaarel Kaljurand\inst{1} \and Tobias Kuhn\inst{1,2}}
%
\authorrunning{Kaarel Kaljurand et al.}
%
\tocauthor{Kaarel Kaljurand, and Author Name}
\institute{Institute of Computational Linguistics, University of Zurich,
Switzerland
\and
Chair of Sociology, in particular of Modeling and Simulation, ETH Zurich, Switzerland
\smallskip\\
\email{kaljurand@gmail.com}, \email{kuhntobias@gmail.com}}

\maketitle 

\begin{abstract}
We describe a semantic wiki system with an underlying controlled natural
language grammar implemented in Grammatical Framework (GF).
The grammar restricts the wiki content to a well-defined
subset of Attempto Controlled English (ACE), and facilitates a precise
bidirectional automatic translation between ACE
and language fragments of a number of other natural languages,
making the wiki content accessible multilingually.
Additionally, our approach allows for automatic translation into the Web
Ontology Language (OWL), which enables automatic reasoning over the wiki content.
The developed wiki environment thus allows users to build, query and view
OWL knowledge bases via a user-friendly multilingual natural language interface.
As a further feature,
the underlying multilingual grammar is integrated into the wiki
and can be collaboratively edited to extend the vocabulary of the wiki or even
customize its sentence structures.
This work demonstrates the combination of the existing technologies of
Attempto Controlled English and Grammatical Framework, and is implemented
as an extension of the existing semantic wiki engine AceWiki.
\end{abstract}

\section{Introduction}

Wikis are user-friendly collaborative environments for building knowledge bases
in natural language.
The most well-known example is Wikipedia, an encyclopedia that is
being built by around 100,000 users in hundreds of different languages, with numerous
other wikis for smaller domains.
Semantic wikis \cite{bry2012semantic}
combine the main properties of wikis (ease of use, read-write, collaboration,
linking) with
knowledge engineering technology (structured content, knowledge models in the
form of ontologies, automatic reasoning). Semantic wiki editors simultaneously
work with the natural language content and its underlying formal semantics
representation. The resulting wikis offer more powerful
content management functions, e.g. dynamically created pages based on semantic
queries and detection of semantic errors in the content,
but have to somehow meet the challenge of keeping the user interface as simple
as expected from wikis.
The existing semantic wiki engines (e.g.
Semantic Mediawiki\footnote{\url{http://semantic-mediawiki.org/}},
Freebase\footnote{\url{http://www.freebase.com/}}) support the inclusion of
semantics in the form of RDF-like subject-predicate-object triples,
e.g. typed wikilinks (predicates)
between two articles (the subject and the object).

Our approach to semantic wikis is based on controlled natural language (CNL) \cite{short_wyner2009cnlmain}.
A CNL is a restricted version of a natural language. For CNLs like
Attempto Controlled English (ACE) \cite{short_fuchs:reasoningweb2008},
the syntax is precisely defined,
the sentences have a formal (executable) meaning,
and they come with end-user documentation describing syntax,
semantics and usage patterns.
CNLs and their editing tools support the creation of texts that are
natural yet semantically precise,
and can thus function well in human-machine communication.
CNL-based wikis --- such as AceWiki \cite{kuhn:swui2008}, on which our approach is based
--- can offer greater semantic
expressivity compared to traditional semantic wikis (e.g. OWL instead of RDF),
but their user interface is easier to work with
(because it is still based on natural language).

In this paper we describe a semantic wiki system with an underlying controlled
natural language grammar implemented in Grammatical Framework (GF).
The grammar restricts the wiki editors into a well-defined
subset of ACE that is
automatically translatable into the Web Ontology Language (OWL)
\cite{OWL_2_Web_Ontology_Language_Document_Overview_Second_Edition}
and thus enables
automatic semantic reasoning over the wiki content. Additionally, the grammar
facilitates a precise bidirectional automatic translation between ACE
and language fragments of a number of other natural languages.
The developed wiki environment thus allows users to build, query and view
OWL knowledge bases via a user-friendly multilingual natural language interface.
The underlying multilingual grammar is integrated into the wiki itself
and can be collaboratively edited to extend the vocabulary and even
customize the multilingual representations of ACE sentences.
Our work demonstrates the combination of the existing technologies of
ACE and GF, and is implemented by extending
the existing ACE-based semantic wiki engine AceWiki with
support to multilinguality and collaborative GF grammar editing.
The main goal of this work is to explore natural language grammar based
semantic wikis in the multilingual setting. As a subgoal we ported
a fragment of ACE to several natural languages
(other than English)
in a principled way by implementing a shared abstract syntax.
The wiki environment allows us to test the usefulness of this work and
furthermore collaboratively improve the initial ports.
The overall work is part of the the EU research project
MOLTO\footnote{\url{http://www.molto-project.eu}}.

This paper is structured as follows:
in Section \ref{section:Related_work} we review related work;
in Section \ref{section:Underlying_technologies} we introduce the core features
of the existing tools and technologies employed in the rest of the paper
(namely ACE, GF and AceWiki);
in Section \ref{section:Multilingual_ACE} we describe the multilingual
GF-implementation of ACE;
in Section \ref{section:AceWiki-GF} we discuss the extension of AceWiki
based on the GF-implementation of ACE;
in Section \ref{section:Evaluation} we provide an initial evaluation of our
system;
in Section \ref{section:Future_work} we summarize our main results and
outline future work.

\section{Related work}
\label{section:Related_work}

The related work falls into several categories such as
multilingual CNLs, CNL-based wikis, multilingual wikis, multilingual
ontologies, and ontology verbalization.

Many general purpose and domain-specific controlled natural languages
have been developed based on many different natural languages
\cite{pool:claw2006}. However, there has not been
an effort to bring them under the same semantic model or synchronize their
development in a community-driven manner \cite{luts:cnl2010}.
Our multilingual ACE grammar is an experiment in this direction.
A multilingual version of ACE (in GF) was first investigated in
\cite{ranta:cnl2009_revised}.
Our current implementation is partly an extension of this work. A similar
work is
\cite{gruzitis:phd}, which builds a bidirectional interface between a
controlled fragment of Latvian and OWL, using ACE as an interlingua, and
implementing the interface using GF.

The main CNL-based wiki that we are aware of is AceWiki which is also the basis
of our work and will be discussed below.
\cite{ghidini2012modeling} describes the MoKi semantic wiki engine which
offers a ``lightly-structured access mode'' for its structured content (OWL).
In this mode the content is displayed as an uneditable ACE text;
editing is supported for the simpler \emph{isA} and \emph{partOf} statements
using templates that combine CNL with HTML-forms, or using a native OWL
syntax.
In terms of multilinguality our wiki system has some similarities
with the OWL ontology editor described in \cite{bao:acita2009} which allows the
user to view the ontology in three CNLs, two based on English and one on
Chinese.
As the main difference compared to these systems,
our system uses the CNLs as the only user interface for both editing and
viewing.

The research on GF has not yet focused on a wiki-like tool built on top of
a GF-based grammar or application. Tool support exists mostly for users
constructing single sentences (not texts) and working alone
(not in collaboration).
A notable exception is \cite{meza:gotal2008}, which investigates using GF
in a multilingual wiki context, to write restaurant reviews on the
abstract language-independent level by constructing GF abstract trees.

Even though the mainstream wiki engines generally allow for the wiki articles
to be written
in multiple languages, these different language versions exist independently
of each other and only article-level granularity is offered by the system for
interlinking the multilingual content. Some recent work targets that problem
though, e.g. the EU project CoSyne\footnote{\url{http://www.cosyne.eu/}}
develops a technology for the multilingual content synchronization in wikis
by using machine translation.

Ontology languages (such as RDF, OWL and SKOS) typically support
language-specific labels as attachments to ontological entities (such as classes
and properties). Although the ontological axioms can thus be presented
multilingually, their keywords (e.g. \owl{SubClassOf}, \owl{some}, \owl{only})
are still in English and their syntactic structure is not customizable.
This is clearly insufficient for true ontology verbalization,
especially for expressive ontology languages like OWL as
argued in \cite{cimiano2011lexinfo}, which describes a sophisticated lexical
annotation ontology to be attached to the domain ontology as linguistic
knowledge. Our work can also be seen as attaching (multilingual)
linguistic knowledge to a semantic web ontology.
\cite{davis:cnl2012} discusses a multilingual CNL-based verbalization of
business rules. It is similar to our approach by being implemented in GF but
differs by not using OWL as the ontology language.


\section{Underlying technologies}
\label{section:Underlying_technologies}

\subsection{Attempto Controlled English}
\label{subsection:Attempto_Controlled_English}

Attempto Controlled English (ACE) \cite{short_fuchs:reasoningweb2008} is a general purpose
CNL based on first-order logic.
ACE can be viewed as both a natural language understandable to every
English speaker, as well as a formal language with a precisely defined
syntax and semantics understandable to automatic theorem proving software.
ACE offers many language constructs, the most
important of which are
countable and mass nouns (e.g. `man', `water');
proper names (`John');
generalized quantifiers (`at least 2');
indefinite pronouns (`somebody');
intransitive, transitive and ditransitive verbs (`sleep', `like', `give');
negation, conjunction and disjunction of
noun phrases, verb phrases, relative clauses and sentences;
and anaphoric references to noun phrases through
definite noun phrases, pronouns, and variables.
Texts built from these units are deterministically interpreted
via Discourse Representation Structures (DRS) \cite{kamp:drt1993}, which can
be further mapped to formats supported by existing
automatic reasoners (e.g. OWL, SWRL, FOL, TPTP).
The ACE sentence structures and their
unambiguous interpretations are explained in the end-user documentation
in the form of \emph{construction} and \emph{interpretation} rules.

The grammar of ACE and its mapping to DRS cannot be modified by the end-users
but they can customize ACE in their applications
by specifying a content word lexicon
of nouns, verbs, adjectives, adverbs and prepositions and their mapping to
logical atoms.

While originally designed for software specifications, in the recent years
ACE has been developed with the languages and applications of the
Semantic Web in mind. \cite{kaljurand:phd} describes ACE fragments suitable
for mapping to and from languages like OWL, SWRL and DL-Query.
ACE View \cite{kaljurand:owled2008} and AceWiki are
ACE-based tools for building OWL ontologies.
The study described in \cite{kuhn2013swj} provides evidence that ACE is a
user-friendly language for specifying OWL ontologies, providing
a syntax that is easier to understand and use compared to the standard OWL syntaxes.

\subsection{Grammatical Framework}
\label{subsection:Grammatical_Framework}

Grammatical Framework (GF) \cite{ranta:book2011}
is a functional programming language for building multilingual grammar
applications.
Every GF program consists of an \emph{abstract syntax} (a set of functions
and their categories) and a set of one or more
\emph{concrete syntaxes} which describe how the abstract
functions and categories are linearized (turned into surface strings) in each
respective concrete language. The resulting grammar
describes a mapping between concrete language strings and
their corresponding abstract trees (structures of function names).
This mapping is bidirectional
---
strings can be \emph{parsed} to trees, and trees \emph{linearized} to strings.
As an abstract syntax can have multiple corresponding concrete syntaxes,
the respective
languages can be automatically \emph{translated} from one to the other by
first parsing a string into a tree and then linearizing the obtained tree
into a new string.

While GF can be used to build parsers and generators for formal languages, it
is optimized to handle natural language features like morphological
variation, agreement, and long-distance dependencies. Additionally,
the GF infrastructure provides a \emph{resource grammar library} (RGL),
a reusable grammar library of the main syntactic structures and morphological
paradigms currently covering about 30 natural languages \cite{ranta:lilt2009}.
As the library is accessible via a language-independent API, building
multilingual applications remains simple even if the programmers lack detailed
knowledge of the linguistic aspects of the involved languages.
These features make GF a good framework for the implementation of
CNLs, especially in the multilingual setting \cite{ranta:cnl2010_revised}.
The development of GF has focused on parsing tools, grammar editors, and
extending the grammar library to new languages.
The current algorithm for parsing GF grammars is based on
Parallel Multiple Context-Free Grammars and allows for incremental
parsing, which enables look-ahead editing \cite{angelov2011mechanics}.

\subsection{AceWiki}
\label{subsection:AceWiki}

AceWiki\footnote{\url{http://attempto.ifi.uzh.ch/acewiki/}}
\cite{kuhn2010doctoralthesis} is a CNL-based semantic wiki engine,
implemented in Java using the Echo Web Framework\footnote{\url{http://echo.nextapp.com/}}.
It uses ACE as the content language and OWL
as its underlying semantic framework integrating its main reasoning tasks
(consistency checking, classification and query answering) and making them
available via the ACE-based interface.

The content of an AceWiki instance is written in a subset of ACE formally
defined in a grammar notation called Codeco \cite{kuhn2012jlli}.
The grammar targets an OWL-compatible fragment of ACE, i.e. ACE sentences that
are semantically outside of the OWL expressivity cannot be expressed in the
wiki. This guarantees that all of the AceWiki content can be
automatically translated to OWL in the background.
Additionally, the grammar is used to drive a look-ahead editor
which guides the input of a new sentence by proposing only syntactically
legal continuations of the sentence.

The AceWiki content is structured into a set of articles, each
article containing a sequence of entries which are either declarative
sentences (corresponding to OWL axioms) or questions
(corresponding to OWL class expressions). Additionally informal comments
are supported.
Upon every change in the wiki, an OWL reasoner determines its effect and
possibly flags inconsistencies or updates the dynamically generated parts of
the wiki (e.g. concept hierarchies and answers to questions).

The content words (proper names, nouns,
transitive verbs, relational nouns and transitive adjectives) in the wiki
sentences
map one-to-one (i.e. link) to wiki articles.
Semantically, content words correspond to OWL entities:
proper names to OWL individuals,
nouns to OWL classes, and
the relational words to OWL properties.


\section{Multilingual ACE}
\label{section:Multilingual_ACE}

In order to provide a multilingual interface to AceWiki, we implemented the
syntax of ACE in GF and ported it via the RGL API to multiple natural language
fragments.
(See the
ACE-in-GF website\footnote{\url{http://github.com/Attempto/ACE-in-GF}}
and \cite{camilleri:molto:d11-1} for more details of this work.)
On the ACE side, the grammar implements the subset supported by the AceWiki
Codeco grammar and can be thus automatically tested against the Codeco
implementation to verify the coverage and precision properties.
The implementation accesses the GF English resource grammar through the
language-independent API (Figure \ref{fig:grammar}).
This API makes it easy to plug in other RGL-supported languages. Our current
implementation targets 15 European languages. Most of them
provide full coverage of the ACE syntactic structures, for some languages
a few structures (e.g. verb phrase coordination, some forms of questions)
have not been implemented yet.

\begin{figure}[t]
\centering
\scriptsize
\begin{quote}
\begin{verbatim}
-- ACE noun phrase uses the RGL noun phrase structure
lincat NP = Syntax.NP ;
...
-- noun phrase with the determiner 'every' e.g. 'every country'
lin  everyNP = Syntax.mkNP every_Det ;
...
-- verb phrase with a passive transitive verb and a noun phrase
-- e.g. 'bordered by Germany'
lin v2_byVP v2 np = mkVP (passiveVP v2) (Syntax.mkAdv by8agent_Prep np) ;
\end{verbatim}
\end{quote}
\normalsize
\caption[Fragment of the GF grammar for ACE]
{Fragment of a GF grammar for ACE listing the linearization rules for
the functions \gf{everyNP} and \gf{v2\_byVP}.
There are around 100 such rules.
This GF module (functor) implements
the ACE sentence structures via RGL's API calls
(e.g. \gf{every\_Det},
\gf{mkVP}). A concrete
language implementation parametrizes this functor with a concrete
language resource (English in case of ACE) and possibly overrides
some of the rules with language-specific structures.
For the function categories, the grammar uses categories that are
also used in the ACE user-level documentation, e.g. noun (\gf{N}),
transitive verb (\gf{V2}), noun phrase (\gf{NP}), relative clause.}
\label{fig:grammar}
\end{figure}

While most of the multilingual grammar can be written in a language-neutral
format, the lexicon modules are language dependent. Different languages
have different morphological complexity, e.g. while the Codeco-defined AceWiki
subset of ACE operates with two noun forms (singular and plural) and three
verb forms (infinitive, 3rd person singular and past participle), other
languages (e.g. Finnish) might need many more forms to be used in the various
ACE sentence structures. Fortunately, we can make use
of the RGL calls, e.g.
\gf{mkN} (``make noun'') and
\gf{mkV2} (``make transitive verb''), to create the necessary
(language-specific) lexicon
structures from a small number of input arguments (often just the lemma
form), using the so called smart paradigms \cite{ranta:lilt2009}.

\begin{figure}[ht]
\centering
\includegraphics[width=0.6\textwidth]{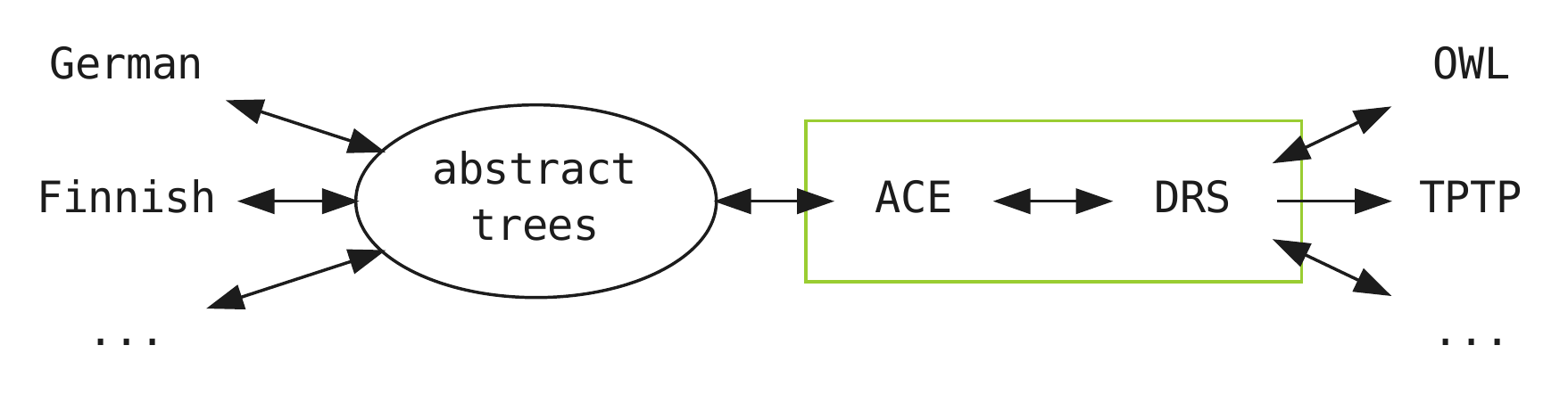}
\caption[Languages]
{Bidirectional mapping between a formal language like OWL and a natural
language like Finnish facilitated by the multilingual GF-implementation of
ACE and various mappings between ACE and other formal languages.}
\label{fig:languages}
\end{figure}

In order to view an ACE text in another language, one needs to parse it
to an abstract tree which can then be linearized into this language
(Figure \ref{fig:lins}). This makes
it possible to map various natural language fragments to the formal languages
that are supported by ACE (e.g. OWL and TPTP) and verbalize such formal
languages via ACE (if this is supported) into various natural language fragments
(Figure \ref{fig:languages}).
For example, the OWL-to-ACE verbalizer
\cite{kaljurand:phd} can be used as a component in a tool that makes
an OWL ontology viewable in a natural language, say Finnish. This tool must
contain a lexicon, i.e. a mapping of each OWL entity to the Finnish word
that corresponds to the ACE category that the verbalizer assigns to the
OWL entity.


\begin{figure}[ht]
\centering
\scriptsize
\begin{quote}
\begin{verbatim}
if_thenS
  (vpS
    (termNP X_Var) (v2VP contain_V2 (termNP Y_Var)))
  (neg_vpS
    (termNP Y_Var) (v2VP contain_V2 (termNP X_Var)))

ACE:     if X contains Y then Y does not contain X
Dutch:   als X Y bevat , dan bevat Y niet X
Finnish: jos X sisältää Y:n niin Y ei sisällä X:ää
German:  wenn X Y enthält , dann enthält Y X nicht
Spanish: si X contiene Y entonces Y no contiene X
\end{verbatim}
\end{quote}
\normalsize
\caption[Abstract tree and its linearizations]
{Abstract tree and its linearizations into five languages which
express the OWL asymmetric property axiom, which
is assigned by the ACE-to-OWL mapping to the ACE sentence.
The linearizations feature different word orders depending on the language.
The tree abstracts away from linguistic features like word order, case,
and gender, although it still operates with syntactic notions such as \emph{negated
verb phrase}.}
\label{fig:lins}
\end{figure}

While the ACE concrete syntax is designed to be unambiguous, i.e.
every supported sentence generates just a single abstract tree, the grammar
in general does not guarantee this property for the other implemented languages.
In some cases it seems to be better to let a user work with an ambiguous
representation if it offers a simpler syntax and if the ambiguity can be
always explained (e.g. via the ACE representation)
or removed in the actual usage scenario (e.g. in a collaborative wiki
environment).


\section{AceWiki-GF}
\label{section:AceWiki-GF}

\begin{figure}[ht]
\centering
\includegraphics[width=0.95\textwidth]{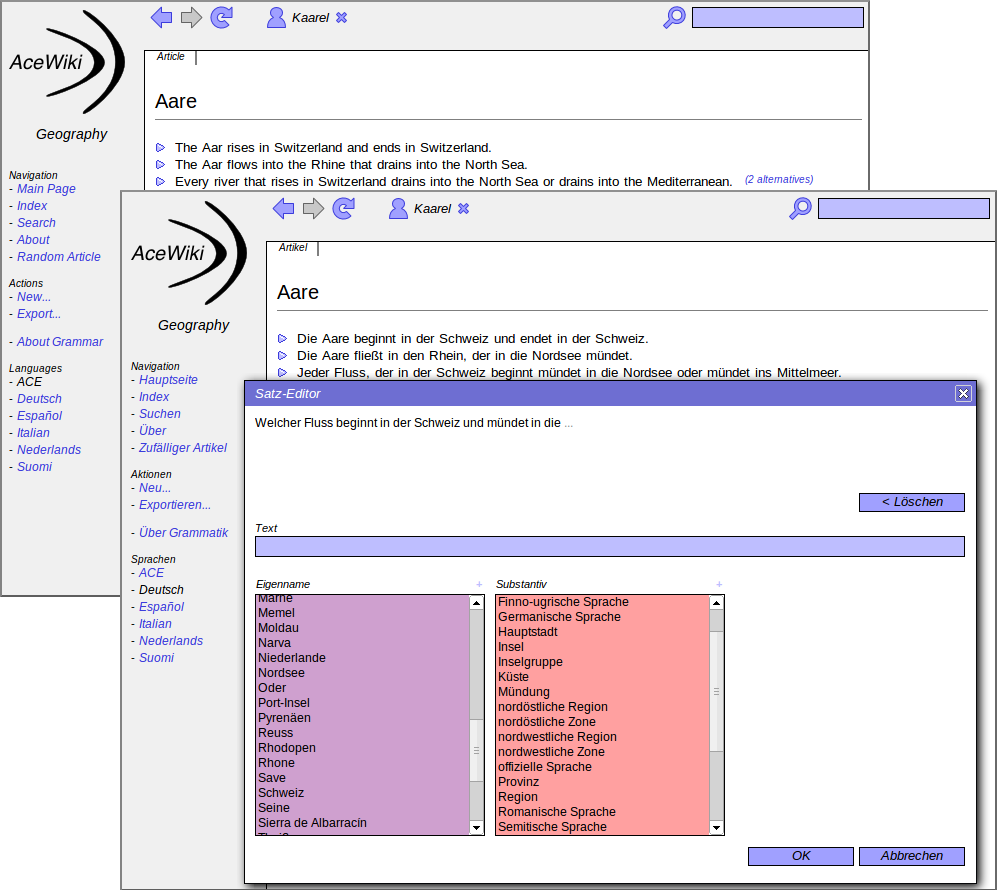}
\caption[Multilingual wiki article]
{Multilingual geography article displayed in ACE and German.
The wiki language (of both the content and the user interface)
can be changed in the left sidebar.
Otherwise the user interface is the same as in AceWiki,
with the look-ahead editor that helps to input syntactically
controlled sentences, in this case offering proper names and common nouns
as possible continuations.}
\label{fig:acewikigf_preditor}
\end{figure}

Our multilingual semantic wiki based on ACE and GF has been realized
as an extension of AceWiki, and is thus (preliminarily) called AceWiki-GF.
Extending AceWiki has allowed us to reuse its infrastructure
(such as look-ahead editing, access to OWL reasoners, the presentation
of reasoning results, and document navigation).
In the following we only describe the main differences and extensions.
(See Section \ref{subsection:AceWiki} for the general discussion of the
AceWiki engine.)

Because AceWiki is a monolingual engine, several modifications had to be
done to accommodate multilinguality, i.e. to support viewing/editing
in multiple languages depending on the users' preferences:

\begin{itemize}
\item the Codeco grammar/parser for ACE was replaced by the
GF-implemented multilingual ACE grammar and a GF parser;
\item the English-specific lexicon editor was replaced by a simple
GF source editor which can be used to edit any GF grammar modules, among them
lexicon modules;
\item the atomic wiki entry, which for the monolingual AceWiki was an ACE
sentence, was changed to a GF abstract tree set. The new representation
is language-neutral and can furthermore represent ambiguity, as explained
in Section \ref{section:Multilingual_ACE};
\item the notion of wiki article/page was extended to also include
arbitrarily named pages (in AceWiki all pages are named by their corresponding
OWL entity) and pages that represent editable grammar modules.
\end{itemize}

The existing AceWiki user interface has largely been preserved; the main
additions are the disambiguation dialog and a menu for setting the
content language, which also determines the user interface language
(Figure \ref{fig:acewikigf_preditor}).
The wiki still follows the main principle of
CNL-based wikis, i.e. that formal notations are hidden. In our case the
user does not see the trees which actually define the content
but only interacts with the natural language sentences.
(Experienced users can still look at the GF parser output providing
information on syntax trees, translation alignment diagrams, etc.)

\subsection{Structure and linking}

In general,
AceWiki-GF follows the AceWiki structure --- the wiki is a set of articles,
each containing a sequence of sentences. New
is the fact that also the grammar definition is part of the wiki and
can be referenced from the articles using wikilinks.

A GF grammar is structured in a way that is naturally representable
as a set of wiki articles. Each grammar module can be stored as a wiki article
and linked to the modules that it imports. Furthermore, grammar modules
have internal structure --- sets of categories and functions (which reference
categories) --- which can be linked to wiki content because the content
is represented as a set of trees (i.e. structures of function names).
One of the benefits of having a grammar definition as part of the
wiki is that it provides an integrated
documentation of the language that the wiki users are required to use.
Note that the full grammar contains also modules which are part of the general
RGL and thus not editable and also not part of the wiki. This resource is
made accessible via external links to the online
RGL browser\footnote{\url{http://www.grammaticalframework.org/lib/doc/browse/}}.

\subsection{Sentence editing}

The user interface for adding and modifying wiki entries is the same as
in AceWiki, i.e. based on sentences and supporting the completion of a
syntactically correct sentence by displaying a list of syntactically legal
words that can follow the partially completed sentence.
The language of the sentence depends
on the chosen wiki language. In case an entry is ambiguous
(i.e. parsing results in multiple trees) then the ambiguity is preserved.
If viewed in another language, multiple different sentences can then occur
as linearizations of the ambiguity. This allows the wiki users who work
via the other language to resolve the ambiguity.
A monolingual way to deal with ambiguity is to implement for every concrete
syntax an additional ``disambiguation syntax'' \cite{ranta:cnl2010_revised},
that overrides the linearizations of the ambiguous constructs to have an
unambiguous, although possibly a more formal-looking notation.
This syntax could be used to display the entry in the case of ambiguity.

We note that some syntax-aware editors, e.g. the GF Syntax
Editor\footnote{\url{http://cloud.grammaticalframework.org/syntax-editor/editor.html}} or the
OWL Simplified English editor \cite{power:cnl2012}, operate more on the
abstract tree level and thus avoid the problem of ambiguous entries. These
approaches also simplify smaller edits e.g. replacing a word in the beginning
of the sentence.
The fact that they abstract away from linguistic details like case and gender
might make them preferable for users with only basic knowledge of the
underlying language.
It is therefore worth exploring these editing approaches also
in the AceWiki-GF context.


\subsection{Lexicon and grammar editing}

\begin{figure}[t]
\centering
\scriptsize
\begin{quote}
\begin{verbatim}
Danish:  country_N = mkN "land" "landet" ;
Dutch:   country_N = mkN "land" neuter ;
Finnish: country_N = mkN "maa" ;
French:  country_N = mkN "pays" masculine ;
German:  country_N = mkN "Land" "Länder" neuter ;
Italian: country_N = mkN "paese" ;
Swedish: country_N = mkN "land" "landet" "länder" "länderna" ;
\end{verbatim}
\end{quote}
\normalsize
\caption[Lexicon entries in the multilingual AceWiki]
{Entries in the multilingual lexicon.
Smart paradigms like \gf{mkN} are used to create the
internal structure of the entry. In many cases giving only the lemma form
to the word class operator is sufficient to get a correct internal structure.
In some cases further forms or information about gender (in some languages)
needs to be added. This makes the user interface to the lexicon relatively
simple and homogeneous across languages.}
\label{fig:acewikigf_grammar}
\end{figure}


Our wiki makes the grammar available as a set of interlinked grammar modules
falling into the following categories:

\begin{itemize}
\item ACE resource grammar (about 30 modules which are typically
identical to their English resource grammar counterparts, sometimes
overriding certain structures);
\item ACE application grammar, reflecting the AceWiki subset of ACE
(one module);
\item instantiation of this grammar for each supported language with
additional modules that describe language-specific overriding of some
of the functions;
\item content word lexicon module(s) for each language.
\end{itemize}

In order to
add a new word to the wiki, a line needs to be added to the lexicon wiki page,
i.e. the page that corresponds to the lexicon module
(Figure \ref{fig:acewikigf_grammar}).
Although editing the lexicon technically means editing the GF grammar,
the lexicon module is conceptually much simpler than the
general grammar module and maps
one-to-one to the respective ACE lexicon structure (for English). The structure
of lexicons in all the supported languages is roughly the same even if some
languages are morphologically more complex (e.g. have more case endings).
The language-specific lexical structures are hidden from
the user behind language-neutral categories like \gf{N} and \gf{V2}
and constructed by functions like
\gf{mkN} and \gf{mkV2}
which are capable of determining the full word paradigm on the basis of only one or
two input forms. Thus, support for multilinguality does not increase
the conceptual complexity of the wiki.

Wiki users experienced in GF are also able to modify the full grammar, although
we do not see many compelling use cases for that as
ACE itself is pre-defined and thus
changing its grammar should not be allowed (e.g. it would break the
functioning of the mapping to OWL). Its verbalization to other languages,
however, is sometimes a matter of taste, and could be therefore
made changeable by the wiki users, e.g. users
can add an alternative formulation of an ACE sentence in some language
by using a GF variant.
Also, the possibility to define arbitrary GF operators can make certain lexicon
entry tasks more convenient.

A change to the underlying grammar (even if only in the lexicon module)
can have the following consequences for the content:
(1) removing a function can render some of the wiki entries (the
ones whose trees use this function) invalid, the user must then reformulate
the respective sentences to conform to the new grammar/lexicon;
(2) altering the linearization of a function might cause some sentences
to become unparsable or ambiguous in the corresponding language.
This does not have an immediate effect on the stored wiki content because
the storage is based on trees, but if an existing sentence is submitted
again to the parser then it might fail or result in more trees than before.
A general change to a grammar module (e.g. removing a category)
can also make the whole grammar invalid, which obviously should be avoided.

\subsection{Underlying semantic representation}

As in the original AceWiki, each AceWiki-GF entry has a corresponding OWL
representation. It is obtained by linearizing the
abstract tree of the entry as an ACE sentence (using the multilingual grammar)
and then translating it
to OWL (as defined in \cite{kaljurand:phd}).
In the ACE representation each content word is
annotated with its word class information and the corresponding OWL entity,
which is currently derived from the lemma form of the ACE word.
Ambiguous wiki entries map in general to multiple OWL forms
(although this is not necessarily the case). Such entries are not
included in the underlying semantic representation.


\subsection{Multilinguality}

The content of our wiki can be currently made available in up to 15 languages,
which form a subset of the RGL that has been tested in the context
of the multilingual ACE grammar.
In principle every RGL language (that exists now or will be added to
the RGL in the future) can be plugged in, because
we access the RGL via its language-neutral API.
However,
language-specific customization of some of the phrase structures is usually
necessary as discussed in Section \ref{section:Multilingual_ACE}.

For a concrete wiki instance a smaller number of languages might
be initially preferred and more translations of the wiki
content could be added gradually.
In addition to the wiki reader and the wiki editor, there is now a third type of
a wiki user, namely the translator. Their main task is to
translate all existing words
by referencing the correct operators in the RGL morphological API
and to check if the automatically generated translations are accurate
with respect to ACE.
The skillset for this task therefore includes the knowledge of ACE and
the RGL morphology API.



\subsection{Implementation}

Apart from having been implemented as an extension of AceWiki, the discussed
wiki engine is supported by two external (and independently developed) tools.
First, the GF Webservice \cite{bringert:eacl2009} provides
linearization and parsing (and the related look-ahead) services
for GF grammars.
The GF Webservice has been recently extended to provide a
GF Cloud Service
API\footnote{\url{http://cloud.grammaticalframework.org/gf-cloud-api.html}}
which additionally allows for modifications to the grammar.
Secondly, the ACE parser APE\footnote{\url{http://github.com/Attempto/APE}}
provides the mapping of ACE sentences to the OWL form
(as is the case also for the monolingual AceWiki).
The current implementation of AceWiki-GF is available on
GitHub\footnote{\url{http://github.com/AceWiki/AceWiki}} and can be used
via some demo wikis\footnote{\url{http://attempto.ifi.uzh.ch/acewiki-gf/}}.

\section{Evaluation}
\label{section:Evaluation}

In previous work, two usability experiments have been performed on AceWiki with altogether 26 participants \cite{semwiki2009_kuhn}. The results showed that AceWiki and its editor component are easy to learn and use. Another study confirmed that writing ACE sentences with the editor is easier and faster than writing other formal languages \cite{kuhnhoefler2012coral}. It has also been demonstrated that ACE is more effective than the OWL Manchester Syntax in terms of understandability \cite{kuhn2013swj}.
As these previous studies did not include the multilinguality features,
the evaluations presented below focus on the multilingual grammar aspects.

We first evaluated how many syntactically correct sentences of the AceWiki ACE
subset the multilingual grammar accepts. To that aim, we used the AceWiki
Codeco testset which is an exhaustive set of sentences with length of up to
10 tokens (19,422 sentences, disregarding some deprecated ACE sentences)
\cite{kuhn2010doctoralthesis}.
The GF-based ACE grammar successfully covers all these sentences.

Next, we measured the syntactic precision by randomly generating large numbers of
sentences at different tree depths and parsing them with both the ACE parser
and the Codeco parser. The precision of the grammar was found to be sufficient
although not perfect. The main deficiency compared to the Codeco grammar is
the lack of DRS-style anaphoric reference modeling. In practice this means that
some accepted sentences will be rejected by the ACE-to-OWL translator on the
grounds of containing unresolvable definite noun phrases.
Ignoring such sentences the precision was 98\% (measured at tree depth of 4
for which the sentence length is 11 tokens on average).

The ambiguity level of ACE sentences (of the Codeco testset) was found
to be 3\%. In these relatively rare cases, involving complex
sentences, the grammar assigns two abstract trees
to an input ACE sentence. This is always semantically harmless ambiguity
(i.e. it would not manifest itself in translations) resulting from the rules
for common nouns and noun phrases which accept similar input structures.
While the coverage and precision are measures applicable only to the ACE grammar
(because an external definition and a reference implementation of ACE exists),
the ambiguity can be measured for all the languages implemented in the grammar
by linearizing trees in a given language and checking
if the result produces additional trees when parsed. Some semantically
severe ambiguities were found using this method (e.g.
occasional subject/object relative clause ambiguity in Dutch and German
triggered by certain combinations of case and gender,
double negation ambiguity in some Romance languages).
These findings can either be treated in the grammar i.e. in the
design of the respective controlled languages or highlighted in the wiki
environment in a way that they can be effectively dealt with.

To measure the translation quality we looked at the translations of 40
ACE sentences using 20 lexicon entries.
The sentences were verbalizations of a wide
variety of OWL axiom structures (also used in \cite{kuhn2013swj}).
We wanted to check whether the meaning in all the languages adheres to
the precise meaning of OWL statements.
The translations covered nine languages
(Catalan, Dutch, Finnish, French, German, Italian, Spanish, Swedish, and Urdu)
and were checked by
native speakers to evaluate the translations with respect to the
original ACE sentence and the ACE interpretation rules
\cite{camilleri:molto:d11-1}. In general, the translations were
found to be acceptable and accurate although several types of errors were
found, mainly caused by the fact that the lexicon
creators were not very familiar with the respective languages. Concretely,
the following four error types were observed:
\begin{description}
\item[RGL Errors.] Some problems (e.g. missing articles in Urdu)
were traced to errors in the resource
grammar library, and not in our ACE application grammar.
\item[Incorrect use of smart paradigms.] Several mistakes were caused by
an incorrect use of the RGL smart paradigms, either by applying a regular
paradigm to an irregular word or simply providing the operator with an
incorrect input (e.g. a genitive form instead of a nominative).
\item[Stylistic issues.] A further problem were stylistic issues, i.e.
structures that are understandable but sound unnatural to a native speaker,
e.g. using an inanimate pronoun to refer to a person.
\item[Negative determiners.] We experienced that translating
sentences with negative determiners such as `no',
e.g. ``every man does not love no woman'' or ``no man does not love
a woman'' can result in meaning shifts between languages.
This was eventually handled by extending the RGL to
include noun phrase polarity.
\end{description}
Most of these problems are easy to fix by a native speaker with GF skills.
(We assume that if there is enough interest in the port of a particular wiki
into another language, it should be possible to find such a person.)
A more conclusive evaluation is planned that includes the
wiki environment and uses a larger real-world vocabulary.

\section{Discussion and future work}
\label{section:Future_work}

The main contribution of our work is the study of CNL-based
knowledge engineering in a semantic wiki environment. The main novelty
with respect to previous work is making the wiki environment multilingual.
As the underlying technologies we have used Attempto Controlled English, which
is a syntactically user-friendly formal language and provides a mapping to the
expressive ontology language OWL, and Grammatical Framework, which was used to
provide a
multilingual bidirectional interface to ACE covering several natural
languages.
We have built the implementation on top of AceWiki, an existing monolingual
semantic wiki engine.
In order to make our system multilingual, the architecture of AceWiki was
generalized. Although the underlying implementation has become more complex,
the user interface has largely remained the same.
On the (multilingual) lexicon editing side, this is mainly due to the support
for smart paradigms that GF provides via its RGL.
In the future,
we plan to use the grammar-based approach also to implement the other
aspects of the wiki, such as multilingual user interface labels (see \cite{meza:gotal2008}).

The current approach generates the OWL representations using the existing
ACE-to-OWL translator. An alternative method is to implement this translator
also in GF. In this way the users would have full control over what
kind of OWL axioms can be generated because they can edit the OWL mapping
(concrete syntax) in the wiki.
The semantic aspects of the wiki could also be generalized to allow for any
kind of ACE-based reasoning, offered by tools like RACE
\cite{fuchs:cnl2010_revised} or TPTP reasoners.

The presented work can be also extended in various more general directions.
Although the current system is ACE-based,
its general architecture allows for any grammar to be used as the basis
of the wiki content as long as it is implemented in GF.
Such alternative
grammars might not map naturally to a language like OWL and are thus
less interesting in the context of the Semantic Web. Examples are grammars
for a tourist phrase book, a museum catalog, a technical manual, or
a collection of mathematics exercises. Such wikis would mainly profit from the
supported multilinguality and not so much from semantic web style reasoning,
or may need other forms of reasoning.

Another direction is to improve the grammar editing features of the environment
and to develop the system into a tool for collaboratively designing CNLs. The wiki
users could take e.g. the ACE grammar as starting point and customize it
for a specific domain, possibly changing some of its original features
and design decisions. The wiki sentences could then serve as unit/regression
test sets to check the currently effective grammar implementation.

\subsubsection{Acknowledgments}
The research leading to these results has received funding from the
European Union's Seventh Framework Programme (FP7/2007-2013)
under grant agreement FP7-ICT-247914.
The authors would like to thank Norbert E. Fuchs
for useful comments on the draft of this paper.

\bibliography{bib}
\bibliographystyle{plain}
\end{document}